\documentclass[runningheads]{llncs}

\usepackage{eccv}

\usepackage{eccvabbrv}

\usepackage{graphicx}
\usepackage{booktabs}
\usepackage{diagbox}
\usepackage[accsupp]{axessibility}  

\usepackage[breaklinks,colorlinks=false,citebordercolor={0 1 0}]{hyperref}
\usepackage{orcidlink}
\usepackage{multirow}
\usepackage[table]{xcolor}
\definecolor{graybg}{rgb}{0.9, 0.9, 0.9}

\begin{document}

\title{Frames2Residual: Spatiotemporal Decoupling for Self-Supervised Video Denoising}

\titlerunning{Spatiotemporal Decoupling for Self-Supervised Video Denoising}

\author{Mingjie Ji\inst{1}\orcidlink{0009-0007-0574-5457} \and
Zhan Shi\inst{1}\orcidlink{0009-0007-7071-0525} \and
Kailai Zhou\inst{2}\orcidlink{0000-0001-5449-3929} \and
Zixuan Fu\inst{2}\orcidlink{0009-0005-5613-3624} \and
Xun Cao\inst{1, 3, }\thanks{Corresponding author.}\orcidlink{0000-0003-3094-4371}}

\authorrunning{M.~Ji et al.}

\institute{
    Nanjing University, Nanjing, China \and 
    Nanyang Technological University, Singapore \and
    Key Laboratory of Optoelectronic Devices and Systems with Extreme Performances of MOE, Nanjing University, Nanjing, China \\
    \email{\{jmj, shizhan\}@smail.nju.edu.cn, \{kailai.zhou, zixuan006\}@ntu.edu.sg, caoxun@nju.edu.cn}
}

\maketitle

\begin{abstract}
Self-supervised video denoising methods typically extend image-based frameworks into the temporal dimension, yet they often struggle to integrate inter-frame temporal consistency with intra-frame spatial specificity. Existing Video Blind-Spot Networks (BSNs) require noise independence by masking the center pixel, this constraint prevents the use of spatial evidence for texture recovery, thereby severing spatiotemporal correlations and causing texture loss. To address this, we propose Frames2Residual (F2R), a spatiotemporal decoupling framework that explicitly divides self-supervised training into two distinct stages: blind temporal consistency modeling and non-blind spatial texture recovery. In Stage 1, a blind temporal estimator learns inter-frame consistency using a frame-wise blind strategy, producing a temporally consistent anchor. In Stage 2, a non-blind spatial refiner leverages this anchor to safely reintroduce the center frame and recover intra-frame high-frequency spatial residuals while preserving temporal stability. Extensive experiments demonstrate that our decoupling strategy allows F2R to outperform existing self-supervised methods on both sRGB and raw video benchmarks. Code is available at \url{https://github.com/m1NGGi/F2R}.
\end{abstract}

\section{Introduction}

\begin{figure*}[t]
  \centering
   \includegraphics[width=\linewidth]{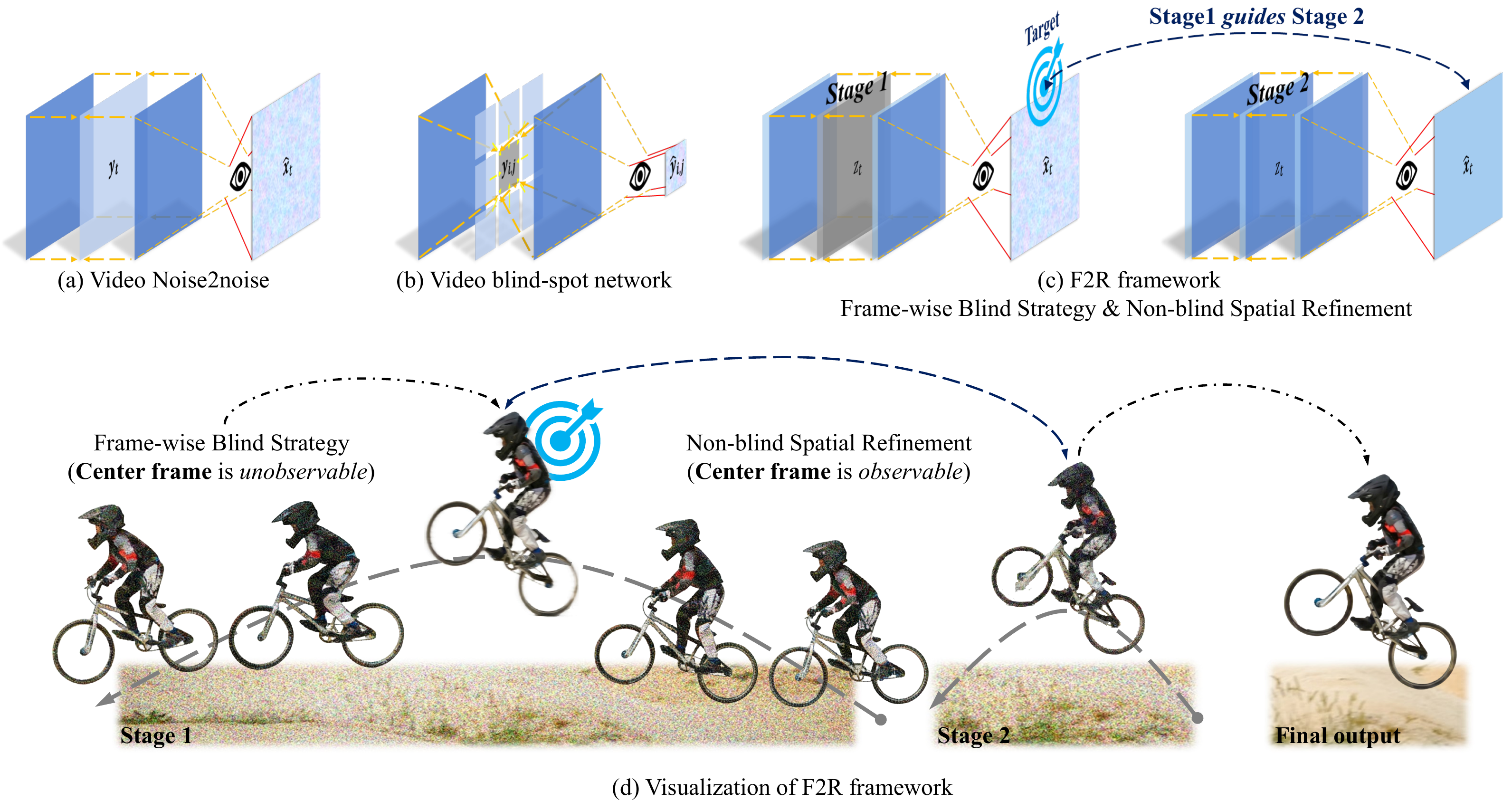} 
  \caption{Comparison of self-supervised video denoising paradigms during the inference phase.
  (a) Video Noise2Noise. Warping-based supervision violates noise independence and causes warping artifacts, leading to blurred details. (b) Video Blind-Spot Network. The inherent imposition of pixel discontinuities severs vital \emph{spatiotemporal correlations}, resulting the loss in texture.
  (c) The proposed F2R framework. F2R employs joint spatial inputs $\{z_i\}$, where $z_i=\mathrm{cat}(\hat{x}_i,r_i)$, $r_i=y_i-\hat{x}_i$, and $\hat{x}_i$ is derived from a pre-trained image denoiser $\mathcal{D}$.
  (d) Visualization of F2R framework. The gray dashed line illustrates the temporal consistency established by Stage~1. The final output acquires spatial specificity while maintaining temporal consistency, thereby effectively restoring \emph{spatiotemporal correlations}.}
  \label{f0}
  \vspace{-0.7cm}
\end{figure*}

Deep learning has achieved remarkable progress in image and video restoration. While image denoising leverages spatial self-similarity to suppress noise \cite{zamir_restormer_2022, huang_neighbor2neighbor_2021}, its efficacy is fundamentally constrained by the spatial entanglement of signal and noise \cite{li2025back}. To address this, video denoising introduces temporal dimension, utilizing correlations across frames to robustly identify coherent structures against random fluctuations \cite{song2022tempformer,vaksman2021patch,tassano2020fastdvdnet}. However, the success of these methods predominantly hinges on supervised learning with paired datasets. Ground Truth (GT) is rarely available in practical applications like live-cell fluorescence microscopy or ultrafast transient imaging \cite{yue2020supervised,sadri2024unsupervised}. This limitation necessitates self-supervised approaches that operate directly on raw noisy videos.

Most existing self-supervised video methods extend image-based frameworks, primarily Noise2Noise (N2N) \cite{lehtinen2018noise2noise} and Blind-Spot Networks (BSNs) \cite{krull2019noise2void,wu2020unpaired} to the temporal dimension. Early Video N2N approaches \cite{davy2019non,dewil2021self,ehret2019model} typically rely on motion compensation to align adjacent frames as supervision targets. Unfortunately, the interpolation inherent in warping inevitably disrupts pixel continuity and noise statistics \cite{chen2025spatiotemporal}, thereby fundamentally violating the independence assumption critical for self-supervision. Consequently, these methods often suffer from severe ghosting artifacts or over-smoothing in dynamic scenes (see Fig.~\ref{f0}(a)). To avoid this issue, Video BSNs frameworks \cite{sheth2021unsupervised} provide a statistically grounded alternative. By approximating the conditional expectation $\mathbb{E}[x_{t,i,j}|\Omega]$ solely from the spatiotemporal neighborhood $\Omega$ (see Fig.~\ref{f0}(b)), they circumvent the ill-posed motion alignment problem, ensuring that the training objective remains pristine and free from warping-induced corruptions. However, this strict exclusion strategy entails a specific trade-off. While this formulation strictly satisfies the noise independence assumption, the exclusion of direct spatial evidence leads to inherent pixel discontinuities, thereby disrupting continuous local variations \cite{wang2022blind2unblind,lee2022ap}. Consequently, this severs vital \emph{spatiotemporal correlations}, defined as the structural continuity between a pixel and its local spatiotemporal neighborhood, ultimately leading to texture loss.

Simultaneously enforcing a blind-spot constraint for noise independence and utilizing direct spatial evidence for texture recovery are fundamentally conflicting objectives. To resolve this inherent conflict, we propose \textbf{Frames2Residual (F2R)}, a two-stage framework that fundamentally decouples temporal consistency modeling from spatial texture restoration. To enable this spatiotemporal decoupling and strictly focus the network on recovering high-frequency spatial textures lost by BSNs, we introduce a residual-domain learning formulation. Specifically, F2R leverages pre-computed image denoising outputs as a structural baseline (illustrated in Fig.~\ref{f0}(c)). By explicitly offloading static structural modeling to this prior, the model can concentrate exclusively on recovering spatial residuals. Similarly, to relieve the network from complex motion modeling during temporal aggregation, we introduce pre-computed optical flow. Directly applying pre-computed flow fields for explicit image warping inevitably corrupts vital details. Instead, we embed them into flow-guided implicit alignment modules, empowering the network to actively suppress resampling artifacts.

Under this residual-domain formulation, F2R utilizes the blindly estimated temporal consistency as a supervision target, creating a safe pathway to reintroduce direct spatial evidence for explicit texture recovery (illustrated in Fig.~\ref{f0}(d)). Specifically, Stage 1 trains a Blind Estimator (BE) via a frame-wise blind strategy. Because the target frame is unobservable, a Flow-Guided Attention Alignment Module (FAAM) replaces standard U-Net skip connections to stabilize cross-frame correlations and extract a temporally consistent anchor. Guided by this anchor, Stage 2 trains a non-blind Spatial Refiner (SR) via a recorruption strategy (substituting the original noisy center frame with a noise-injected anchor) to force active spatial texture recovery. With the target visible, a Flow-Guided Deformable Alignment Module (FDAM) leverages this spatial evidence to correct flow misalignments. By synthesizing complete spatiotemporal features, the SR precisely restores the missing spatial residuals absent in the structural baseline. Ultimately, this decoupled design unifies the inter-frame temporal consistency established in Stage 1 with the intra-frame spatial specificity recovered in Stage 2, successfully restoring the \emph{spatiotemporal correlations}. Extensive evaluations on both sRGB and raw video benchmarks demonstrate that our proposed framework achieves state-of-the-art video denoising performance against current unsupervised methods.

In summary, our main contributions are:
\vspace{-0.2cm}
\begin{itemize}
\item We identify an inherent conflict in self-supervised video denoising: enforcing noise independence through blind-spot prediction prevents the model from exploiting direct spatial evidence for texture recovery. To resolve this issue, we propose F2R, a two-stage framework that decouples temporal consistency modeling from spatial texture restoration.
\item To enable this decoupling, we introduce a residual-domain learning formulation that leverages image denoising priors as a structural baseline, allowing the model to focus on recovering high-frequency spatial residuals while effectively aggregating temporal information.
\item Our approach surpasses current unsupervised methods on both raw and sRGB video denoising benchmarks, demonstrating its superior effectiveness.
\end{itemize}

\section{Related Work}

\subsection{Supervised Video Denoising}
Supervised video denoising typically relies on GT sequences to learn spatiotemporal features \cite{chan2021basicvsr,chan2022basicvsr++,li2022unidirectional,liang2024vrt,liang2022recurrent,maggioni2021efficient,tassano2020fastdvdnet,vaksman2021patch,wang2019edvr,yue2020supervised,xue2019video}. TOFlow \cite{xue2019video} and DVDNet \cite{tassano2019dvdnet} explicitly computed optical flow to align neighboring frames. To avoid artifacts from flow estimation errors, FastDVDNet \cite{tassano2020fastdvdnet} proposed a cascade of U-Nets without explicit motion compensation. EDVR \cite{wang2019edvr} and RViDeNet \cite{yue2020supervised} utilize Deformable Convolution Networks (DCN) \cite{wang2019edvr} for feature-level alignment, while Transformer-based models like VRT \cite{liang2024vrt}, RVRT \cite{liang2022recurrent}, and Tempformer \cite{song2022tempformer} leverage attention mechanisms for long-range temporal modeling. Despite their high performance, the heavy reliance on paired clean-noisy data severely restricts their applicability in real-world scenarios.

\subsection{Self-supervised Video Denoising}
In the absence of GT, self-supervised video denoising has emerged as a vital paradigm to circumvent the dependence on paired training data. Early self-supervised approaches \cite{davy2019non,dewil2021self,ehret2019model,yu2020joint} extend the N2N assumption to video by warping adjacent frames to construct clean targets. However, the interpolation and resampling inherent in warping inevitably alter noise statistics and introduce artifacts, potentially violating the independence assumption required for N2N training. Alternatively, some recent works \cite{lee2021restore,zheng2023unsupervised,fu2024temporal} adopt recorruption or pseudo-impairment schemes to construct self-supervised targets without explicit warping. To better preserve statistical validity, Video BSN approaches \cite{zheng2023unsupervised,wang2023recurrent} employ blind-spot constraints, predicting the center pixel solely from its spatiotemporal context. Based on this, implicit alignment methods like UDVD \cite{sheth2021unsupervised} avoid explicit flow artifacts but rely on frame stacking, leading to limited temporal receptive fields. To improve temporal efficiency, recurrent variants \cite{wang2023recurrent,chen2025spatiotemporal} leverage temporal recurrence to reuse hidden states. However, their sequential processing risks overfitting to training sequences. Crucially, these methods often sacrifice local detail and introduce pixel discontinuities due to strict noise independence assumptions, leading to an inevitable loss of \emph{spatiotemporal correlations} that fundamentally restricts the restoration of fine-grained spatial textures.

\section{Methodology}
To resolve the inherent conflict between enforcing noise independence and utilizing spatial evidence for texture recovery, we propose \textbf{F2R}, a framework centered on a \emph{spatiotemporal decoupling} strategy. Specifically, F2R formulates the training process as a two-stage decomposition problem to effectively address the conflicting objectives inherent in video BSNs. By leveraging pre-computed image denoising outputs to offload static structural modeling, F2R focuses exclusively on recovering the remaining residuals. In Stage 1, a Blind Estimator extracts temporal consistency without using spatial evidence. Subsequently in Stage 2, a Spatial Refiner safely reintegrates spatial evidence, synthesizing the complete spatiotemporal information to restore the high-frequency residuals that the image denoiser missed. Both models in this strategy share an identical standard 4-level U-Net backbone, differing exclusively in their skip connections. Specifically, the Blind Estimator employs FAAM, whereas the Spatial Refiner utilizes FDAM. Notably, this dual-model decoupling is exclusive to training; during inference, only the trained Spatial Refiner is deployed.

\subsection{Stage~1: Blind Temporal Estimating via Frame-wise Blind Strategy}
\begin{figure}[htb]
	\centering
	\includegraphics[width=\linewidth]{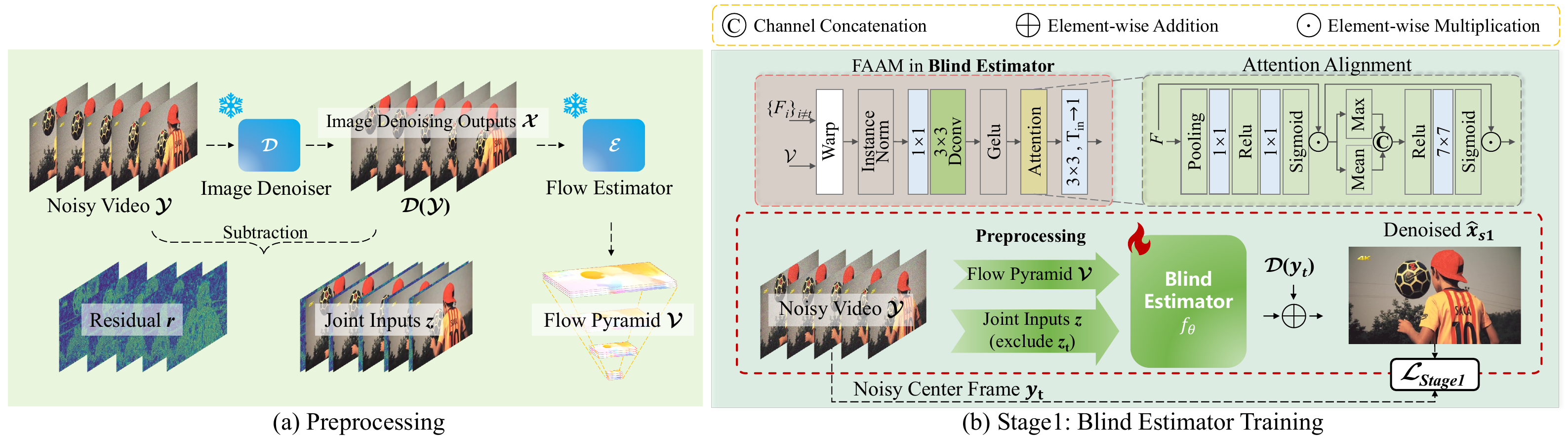}
	\caption{Frame-wise Blind Strategy. (a) Preprocessing for input construction. (b) Blind estimator training. The flow pyramid $\mathcal{V}$ is constructed by downsampling the base flow $\{\mathcal{V}_{i \to t}\}_{i \neq t}$ and halving its magnitude at each level.}
    \label{1}
\end{figure}

Given a noisy video sequence $\mathcal{Y}=\{y_1,\ldots,y_T\}$ where $y_i=x_i+n_i$, the objective of Stage~1 is to isolate the temporal component by constructing a BE tasked with estimating the clean center frame $x_t$ exclusively from inter-frame temporal consistency. To achieve this, we introduce a \textbf{Frame-wise Blind Strategy}, as shown in Fig.~\ref{1}(b). Specifically, we exclude the entire center frame $y_t$ from the input, defining the temporal context as $\Omega_{\mathrm{temporal}} = \{y_i\}_{i \neq t}$. This explicit exclusion naturally satisfies the noise independence assumption, forcing the network to approximate the conditional expectation $\mathbb{E}[x_t|\Omega_{\mathrm{temporal}}]$. By blocking the spatial pathway, we ensure the learned features represent pure temporal consistency, decoupled from intra-frame noise and texture.

\noindent\textbf{Residual-Domain Optimization with Image Denoiser.} Directly regressing the clean frame $x_t$ from temporal neighbors is ill-posed due to complex motions and occlusions. To resolve this, we shift the learning target to the residual domain by leveraging pre-computed image denoising outputs $\mathcal{X}$. A pre-trained image denoiser $\mathcal{D}$ is applied frame-wise to produce the image denoising output $\hat{x}_i=\mathcal{D}(y_i)$. This process yields a deterministic structural baseline $\hat{x}_i$ and a corresponding high-frequency residual $r_i=y_i-\hat{x}_i$. By offloading the modeling of static spatial structures to the image denoiser, this explicit decomposition forces the network to focus exclusively on isolating and recovering the high-frequency texture residual. For each neighboring frame, we construct a joint input $z_i=\mathrm{cat}(\hat{x}_i,r_i)$. The inclusion of $\{\hat{x}_i\}_{i\neq t}$ offers a consistent structural reference to reduce noise ambiguity, while $\{r_i\}_{i\neq t}$ supplies essential high-frequency cues. The process is illustrated in Fig.~\ref{1}(a).

Successfully extracting accurate temporal residuals from these joint inputs requires the network to first overcome the complexities of temporal feature alignment. To prevent the network from expending its capacity on complex motion modeling, we introduce an architectural decoupling to separate spatial extraction from temporal fusion. Specifically, a standard U-Net backbone processes each frame strictly independently to extract intra-frame features, intentionally deferring all inter-frame temporal fusion to the skip connections. Within these connections, we employ pre-computed optical flow to guide implicit feature alignment. This design effectively offloads the alignment burden while empowering the network to actively suppress resampling artifacts inherent to explicit image warping. However, in this blind stage, the absence of the center frame $y_t$ makes aggressive geometric warping ill-posed, as predicting precise offsets without a valid reference often leads to structural distortions.

Therefore, we replace standard U-Net skip connections with a FAAM to implement a conservative filtration strategy. As shown in Fig.~\ref{1}(b), FAAM initially utilizes explicit optical flow $\{\mathcal{V}_{i \to t}\}_{i \neq t}$ (derived from $\mathcal{X}$ using a pre-trained flow estimator $\mathcal{E}$ to mitigate noise interference) to warp neighboring features $\{F_i\}_{i\neq t}$ to the center. To mitigate resulting resampling artifacts, the warped features are stacked and processed through a sequential dual-attention mechanism \cite{hu2018squeeze,woo2018cbam}. A channel attention module acts as a temporal selector to dynamically reweight frames and suppress unreliable motion, followed by a spatial attention module that highlights valid local regions to filter out warping distortions. Finally, a $1\times1$ convolution aggregates these refined features. This distills reliable inter-frame consensus, focusing the BE $f_{\theta}$ on robust temporal residual extraction. The optimization objective is:
\begin{equation}
    \mathcal{L}_{\text{Stage1}} = \mathbb{E} \left\| f_{\theta}(\{z_i\}_{i\neq t}, \mathcal{V}) - r_t \right\|.
\end{equation}
Here, the noisy residual $r_t = y_t - \mathcal{D}(y_t)$ serves as the supervision target, while the network strictly observes only the temporal neighbors $\{z_i\}_{i \neq t}$. The resulting estimate $\hat{x}_{s1}=\hat{x}_t+f_{\theta}(\{z_i\}_{i\neq t},\mathcal{V})$ represents a temporally consistent estimate. However, since this residual is derived exclusively from the decoupled temporal consistency, it inherently lacks frame-specific, non-redundant texture residuals. This intrinsic limitation necessitates a subsequent stage for spatial refinement.

\subsection{Stage~2: Non-blind Spatial Refinement via Recorruption Strategy}
\begin{figure}[htb]
    \centering
    \includegraphics[width=\linewidth]{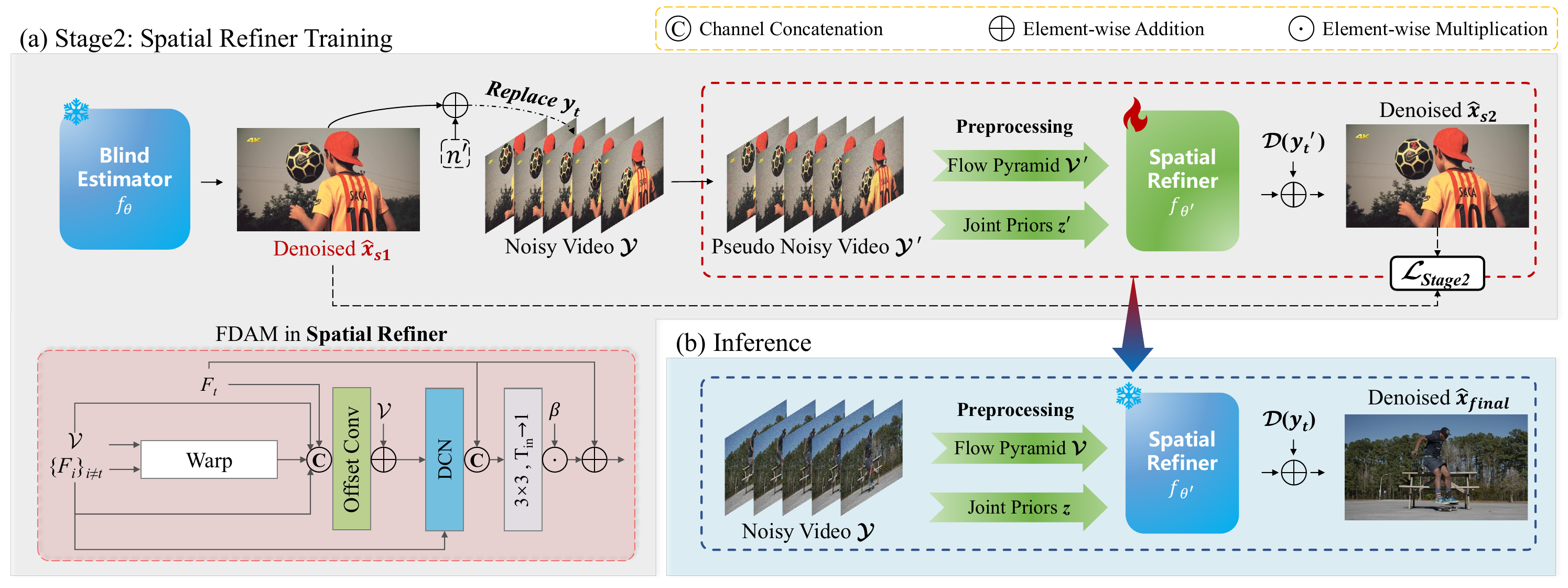}
    \caption{(a) Training and (b) Inference phases of the Spatial Refiner. Notably, the preprocessing operations are strictly identical across both phases. The recorruption noise $n'$ is sampled from the known noise model.}
    \label{fig:stage2}
\end{figure}

Stage 1 successfully isolates temporal consistency, yielding a temporally consistent estimate $\hat{x}_{s1}$. To complete our spatiotemporal decoupling framework, Stage 2 must reintegrate the specific high-frequency spatial textures that the image denoiser $\mathcal{D}$ originally smoothed away, without compromising the temporal stability achieved in Stage 1. To this end, we introduce a \textbf{Non-blind Spatial Refinement} phase driven by a \textbf{Recorruption} strategy (Fig.~\ref{fig:stage2}(a)). This strategy creates a safe pathway to reintroduce direct spatial evidence.

We leverage the fixed mapping of the pre-trained image denoiser $\mathcal{D}$. Since $\mathcal{D}$ systematically suppresses specific high-frequency textures alongside noise, we can train a non-blind SR, $f_{\theta^{\prime}}$, to predict this deterministic spatial gap. We designate $\hat{x}_{s1}$ as our ``temporal anchor''. To construct the training target, we directly synthesize a pseudo noisy sequence $\mathcal{Y}^{\prime}$ by replacing the original noisy center frame $y_t$ in $\mathcal{Y}$ with a recorrupted temporal anchor $y'_t = \hat{x}_{s1} + n^{\prime}$, where $n^{\prime}$ is from a known noise model. The neighboring noisy frames remain strictly unchanged. Injecting this noise $n^{\prime}$ exclusively into the temporal anchor effectively masks the temporal residual embedded in $\hat{x}_{s1}$, compelling the SR to perform active denoising rather than learning a trivial identity mapping.

We reapply the frozen denoiser $\mathcal{D}$ to the recorrupted center frame $y_{t}^{\prime}$. The missing residual is defined as the difference between our temporal anchor and the new image denoising output: 
\begin{equation}
r'_t = \hat{x}_{s1} - \mathcal{D}(y'_t),
\end{equation} 
This precisely quantifies the deterministic texture loss incurred by $\mathcal{D}$.

In stark contrast to Stage 1, the SR $f_{\theta^{\prime}}$ is non-blind and has full access to the unmasked center frame $y_{t}^{\prime}$. The updated joint inputs for the center frame become $z'_t=\mathrm{cat}(\mathcal{D}(y'_t), y'_t-\mathcal{D}(y'_t))$, while neighboring inputs $\{z_i\}_{i\neq t}$ remain identical to Stage 1. This visibility provides the direct spatial evidence necessary for precise calibration of geometric misalignments, crucial for recovering high-frequency spatial textures. Consequently, we employ a FDAM to achieve aggressive, sub-pixel alignment.

FDAM uses DCN \cite{wang2019edvr,fu2024temporal} to handle complex, non-rigid motions missed by pre-computed optical flow. To stabilize DCN optimization, we leverage the explicit optical flow $\{\mathcal{V}^{\prime}_{i \to t}\}_{i \neq t}$ (derived from the updated image denoising sequence $\mathcal{D}(\mathcal{Y}^{\prime})$) as a base offset initialization \cite{chan2022basicvsr++}. The module concatenates the reference feature $F_t$, neighboring features $\{F_i\}_{i\neq t}$, warped neighboring features, and the flow field to regress residual offsets. This decouples global motion from local deformations, focusing DCN on fine-grained sub-pixel adjustments. The aligned features are fused and added back to the reference via a learnable scale parameter $\beta$, ensuring the module can utilize temporal information to actively compensate for the spatial residual extraction process. The non-blind SR is trained to minimize:
\begin{equation}
    \mathcal{L}_{\text{Stage2}} = \mathbb{E} \left\| f_{\theta^{\prime}}(\{z^{\prime}\}, \mathcal{V'}) - r'_t \right\|,
\end{equation}

By minimizing this loss, the SR $f_{\theta^{\prime}}$ learns to fuse the visible intra-frame spatial cues with the broader temporal context, reclaiming the spatial specificity absent from the Stage~1 output. Because $\mathcal{D}$ acts as a consistent spatial filter, the degradation patterns it introduces are identical across both synthetic recorrupted training and actual inference, making the learned residual mapping directly transferable.

As illustrated in Fig.~\ref{fig:stage2}(b), during inference, the BE $f_{\theta}$ is discarded. The converged non-blind SR $f_{\theta^{\prime}}$ is applied directly to the original noisy joint inputs $\{z\}$ and optical flow $\mathcal{V}$ to predict the final residual:
\begin{equation}
     \hat{x}_{\text{final}} = f_{\theta^{\prime}}(\{z\}, \mathcal{V}) + \mathcal{D}(y_t).
\end{equation}

This final output $\hat{x}_{\text{final}}$ seamlessly unifies the inter-frame temporal consistency established in Stage 1 with the intra-frame spatial specificity recovered in Stage 2, successfully reconstructing the vital \emph{spatiotemporal correlations}.

\section{Experiments}
\subsection{Experimental Settings}
\label{Settings}
\noindent \textbf{Datasets.} We evaluate the proposed F2R on two distinct video denoising tasks: synthetic Gaussian denoising and real-world raw video denoising. For synthetic Gaussian denoising, we utilize the DAVIS 2017 \cite{pont20172017} (480P) training set for model training. To generate noisy video sequences, Additive White Gaussian Noise (AWGN) with a standard deviation $\sigma$ varying from 10 to 50 uniformly is added to the clean videos. For evaluation, we employ the DAVIS validation set and the Set8 \cite{tassano2020fastdvdnet} dataset. For real raw video denoising, we employ the CRVD \cite{yue2020supervised} dataset. Due to the limited size of the dataset, we follow the standard unsupervised protocol where all self-supervised models are trained directly on the noisy videos of the CRVD indoor set without accessing ground truth labels.

\noindent \textbf{Implementation Details.} Our framework leverages robust pre-trained priors to ensure both efficiency and performance. For the image denoiser, we employ NAFNet \cite{chen2022simple}, following the training strategy in TAP \cite{fu2024temporal}. For the optical flow estimator, we adopt the pre-trained PWC-Net \cite{sun2018pwc}, following the standard practices in \cite{teed2020raft,dong2024memflow,ranjan2017optical}. During training, we randomly crop patches of size $256 \times 256$ with a batch size of 4. For synthetic Gaussian denoising, the input temporal window size is set to $T=9$, while for real raw video denoising, we set $T=7$. Raw videos are packed into RGBG sequences during training. The feature channel number of the network is fixed at $C=72$, and the network is optimized using the Adam optimizer \cite{kingma2014adam} with $L_1$ loss for synthetic Gaussian denoising and $L_2$ loss for real noise. We train for 200K iterations, with the learning rate initialized at $3 \times 10^{-4}$ and gradually decayed using a cosine annealing strategy \cite{loshchilov2016sgdr}. To ensure statistical consistency in Stage 2, for synthetic experiments, we sample $n'$ from $\mathcal{N}(0, \sigma^2)$ matching the specific noise level $\sigma$ of the current model. In contrast, for real-world tasks where a unified model is trained across all ISO levels, we randomly sample signal-dependent noise based on the calibrated noise profile corresponding to the ISO of the input video. All experiments are implemented in PyTorch and conducted on an NVIDIA L20 GPU. We employ Peak Signal-to-Noise Ratio (PSNR) and Structural Similarity (SSIM) as evaluation metrics.

\begin{figure}[htb]
	\centering
    \vspace{-15pt}
	\includegraphics[width=\linewidth]{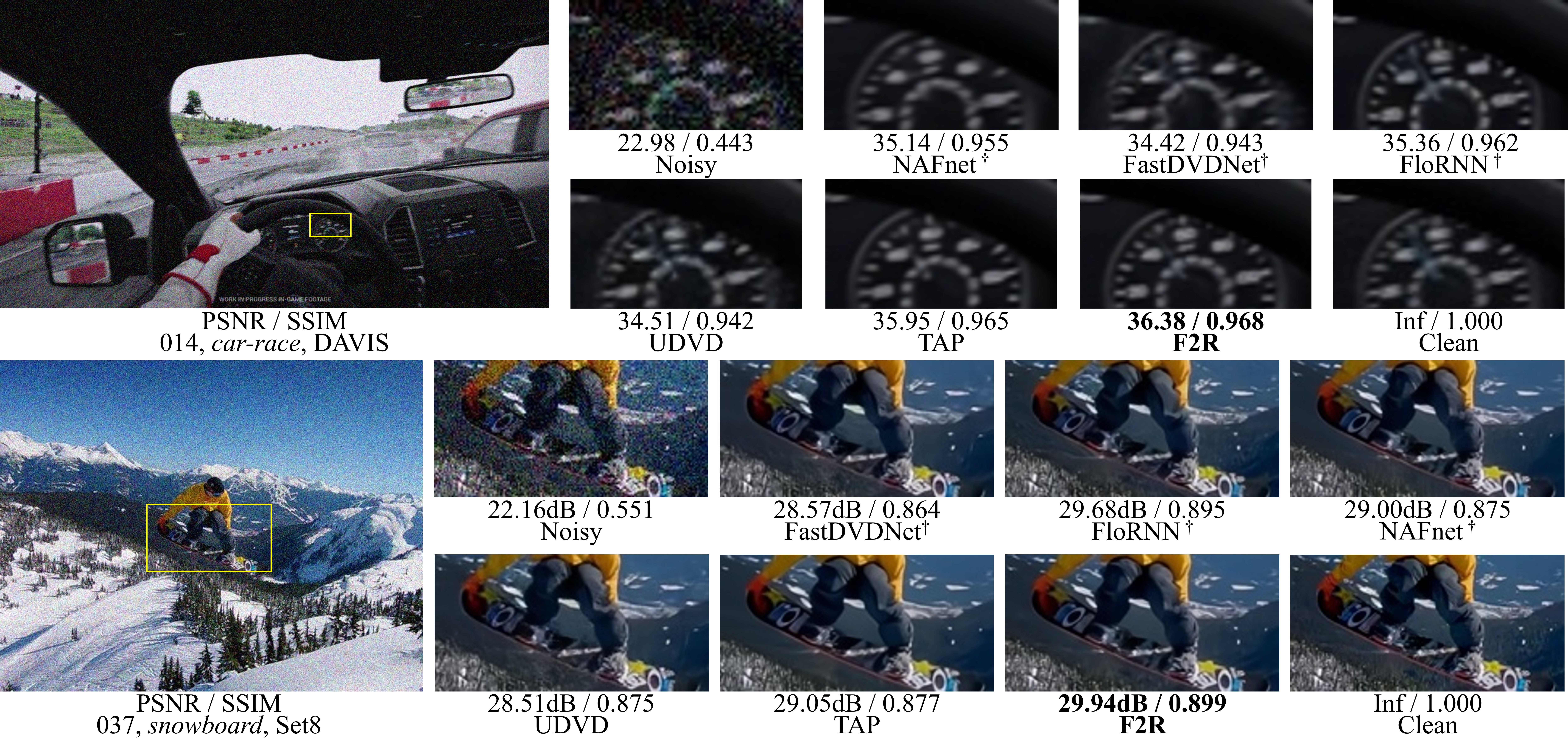}
    \vspace{-20pt}
	\caption{Visual comparison on DAVIS (top) and Set8 (bottom) datasets under noise level $\sigma = 30$. We compare our F2R with supervised (FastDVDNet, FloRNN, NAFNet) and unsupervised (UDVD, TAP) methods, with PSNR/SSIM metrics shown below each patch. $\dag$ indicates the supervised method.}
    \vspace{-25pt}
    \label{f3}
\end{figure}

\subsection{Comparison with State-of-the-Art} 
The experimental results for competing methods are either directly cited from their original papers or reproduced by running the official code. In cases where the code is unavailable, the corresponding results are omitted.  We compare our proposed F2R with state-of-the-art self-supervised video denoising methods, i.e., MF2F \cite{dewil2021self}, RFR \cite{lee2021restore}, UDVD \cite{zheng2023unsupervised}, RDRF \cite{wang2023recurrent}, ER2R \cite{zheng2023unsupervised}, and TAP \cite{fu2024temporal}. For a comprehensive comparison, we also include traditional and supervised video denoising methods such as VBM4D \cite{maggioni2012video}, FastDVDNet \cite{tassano2020fastdvdnet}, PaCNet \cite{vaksman2021patch}, and FloRNN \cite{li2022unidirectional}. Note that methods utilizing the same video sequence for both training and testing are tagged with `T'.

\begin{table*}[htbp]
\centering
\vspace{-20pt}
\renewcommand{\arraystretch}{0.96}
\caption{Quantitative comparison (PSNR/SSIM) on DAVIS and Set8 datasets under different noise levels. The best results for unsupervised methods are \textbf{bolded}, and the second best are \underline{underlined}.}
\label{tab:comparison}
\vspace{-0.2cm}
\resizebox{\textwidth}{!}{%
\begin{tabular}{c|c|l|cc|cc|cc|cc|cc|cc}
\toprule
\multirow{2}{*}{Dataset} & \multirow{2}{*}{ } & \multirow{2}{*}{Method} & \multicolumn{2}{c}{$\sigma = 10$} & \multicolumn{2}{c}{$\sigma = 20$} & \multicolumn{2}{c}{$\sigma = 30$} & \multicolumn{2}{c}{$\sigma = 40$} & \multicolumn{2}{c}{$\sigma = 50$} & \multicolumn{2}{c}{Average} \\
 & & & PSNR & SSIM & PSNR & SSIM & PSNR & SSIM & PSNR & SSIM & PSNR & SSIM & PSNR & SSIM \\
\midrule
\midrule
\multirow{12}{*}{\rotatebox{90}{DAVIS}} 
 & Traditional & VBM4D \cite{maggioni2012video} & 37.58 & - & 33.88 & - & 31.65 & - & 30.05 & - & 28.80 & - & 32.39 & - \\
 \cmidrule{2-15}
 & \multirow{4}{*}{Supervised} 
   & NAFNet \cite{chen2022simple} & 38.79 & 0.965 & 35.37 & 0.933 & 33.47 & 0.904 & 32.17 & 0.879 & 31.18 & 0.858 & 34.20 & 0.908 \\
 & & FastDVD \cite{tassano2020fastdvdnet} & 38.71 & 0.962 & 35.77 & 0.941 & 34.04 & 0.917 & 32.82 & 0.895 & 31.86 & 0.875 & 34.64 & 0.919 \\
 & & PaCNet \cite{vaksman2021patch} & 39.97 & 0.971 & 37.10 & 0.947 & 35.07 & 0.921 & 33.57 & 0.897 & 32.39 & 0.874 & 35.62 & 0.922 \\
 & & FloRNN \cite{li2022unidirectional} & 40.16 & 0.976 & 37.52 & 0.956 & 35.89 & 0.944 & 34.66 & 0.929 & 33.67 & 0.913 & 36.38 & 0.944 \\
 \cmidrule{2-15}
 & \multirow{7}{*}{\begin{tabular}{@{}c@{}}Self-\\supervised\end{tabular}}
   & MF2F-T \cite{dewil2021self} & 38.04 & 0.957 & 35.61 & 0.936 & 33.65 & 0.907 & 31.50 & 0.852 & 29.39 & 0.784 & 33.64 & 0.887 \\
 & & RFR-T \cite{lee2021restore} & 39.31 & - & 36.15 & - & 34.28 & - & 32.92 & - & 31.86 & - & 34.90 & - \\
 & & UDVD \cite{zheng2023unsupervised} & 39.17 & 0.970 & 35.94 & 0.943 & 34.09 & 0.918 & 32.79 & 0.895 & 31.80 & 0.874 & 34.76 & 0.920 \\
 & & RDRF \cite{wang2023recurrent} & 39.54 & 0.972 & 36.40 & 0.947 & 34.55 & 0.925 & 33.23 & 0.903 & 32.20 & 0.883 & 35.18 & 0.926 \\
 & & ER2R-T \cite{zheng2023unsupervised} & 39.52 & - & 36.49 & - & 34.60 & - & 33.29 & - & 32.25 & - & 35.23 & - \\
 & & TAP-T \cite{fu2024temporal} & \underline{39.80} & \underline{0.973} & \underline{36.74} & \underline{0.950} & \underline{34.84} & \underline{0.926} & \underline{33.49} & \underline{0.905} & \underline{32.49} & \underline{0.886} & \underline{35.48} & \underline{0.928} \\
& & \cellcolor{graybg}\textbf{F2R (Ours)} & \cellcolor{graybg}\textbf{40.25} & \cellcolor{graybg}\textbf{0.976} & \cellcolor{graybg}\textbf{37.33} & \cellcolor{graybg}\textbf{0.958} & \cellcolor{graybg}\textbf{35.56} & \cellcolor{graybg}\textbf{0.940} & \cellcolor{graybg}\textbf{34.30} & \cellcolor{graybg}\textbf{0.923} & \cellcolor{graybg}\textbf{33.28} & \cellcolor{graybg}\textbf{0.906} & \cellcolor{graybg}\textbf{36.14} & \cellcolor{graybg}\textbf{0.941} \\
\midrule
\midrule
\multirow{12}{*}{\rotatebox{90}{Set8}} 
 & Traditional & VBM4D \cite{maggioni2012video} & 36.05 & - & 32.19 & - & 30.00 & - & 28.48 & - & 27.33 & - & 30.81 & - \\
 \cmidrule{2-15}
 & \multirow{4}{*}{Supervised} 
   & NAFNet \cite{chen2022simple} & 36.52 & 0.943 & 33.55 & 0.802 & 31.81 & 0.869 & 30.59 & 0.842 & 29.65 & 0.818 & 32.43 & 0.875 \\
 & & FastDVD \cite{tassano2020fastdvdnet} & 36.44 & 0.954 & 33.43 & 0.920 & 31.68 & 0.889 & 30.46 & 0.861 & 29.53 & 0.835 & 32.31 & 0.892 \\
 & & PaCNet \cite{vaksman2021patch} & 37.06 & 0.960 & 33.94 & 0.925 & 32.05 & 0.892 & 30.70 & 0.862 & 29.66 & 0.835 & 32.68 & 0.895 \\
 & & FloRNN \cite{li2022unidirectional} & 37.57 & 0.964 & 34.67 & 0.938 & 32.97 & 0.914 & 31.75 & 0.891 & 30.80 & 0.870 & 33.55 & 0.915 \\
 \cmidrule{2-15}
 & \multirow{7}{*}{\begin{tabular}{@{}c@{}}Self-\\supervised\end{tabular}}
   & MF2F-T \cite{dewil2021self} & 36.01 & 0.938 & 33.79 & 0.912 & 32.20 & 0.883 & 30.64 & 0.841 & 28.90 & 0.778 & 32.31 & 0.870 \\
 & & RFR-T \cite{lee2021restore} & 36.77 & - & 33.64 & - & 31.82 & - & 30.52 & - & 29.50 & - & 32.45 & - \\
 & & UDVD \cite{zheng2023unsupervised} & 36.36 & 0.951 & 33.53 & 0.917 & 31.88 & 0.887 & 30.72 & 0.859 & 29.81 & 0.835 & 32.46 & 0.890 \\
 & & RDRF \cite{wang2023recurrent} & 36.67 & 0.955 & 34.00 & 0.925 & 32.39 & 0.898 & 31.23 & 0.873 & 30.31 & 0.850 & 32.92 & 0.900 \\
 & & ER2R-T \cite{zheng2023unsupervised} & 37.55 & - & 34.34 & - & 32.45 & - & 31.09 & - & 30.05 & - & 33.10 & - \\
  & & TAP-T \cite{fu2024temporal} & \textbf{38.02} & \textbf{0.958} & \underline{35.07} & \underline{0.927} & \underline{33.42} & \underline{0.900} & \underline{32.10} & \underline{0.875} & \underline{31.16} & \underline{0.852} & \underline{33.95} & \underline{0.902} \\
& & \cellcolor{graybg}\textbf{F2R (Ours)} & \cellcolor{graybg}\underline{37.85} & \cellcolor{graybg}\textbf{0.958} & \cellcolor{graybg}\textbf{35.38} & \cellcolor{graybg}\textbf{0.935} & \cellcolor{graybg}\textbf{33.82} & \cellcolor{graybg}\textbf{0.913} & \cellcolor{graybg}\textbf{32.68} & \cellcolor{graybg}\textbf{0.894} & \cellcolor{graybg}\textbf{31.75} & \cellcolor{graybg}\textbf{0.875} & \cellcolor{graybg}\textbf{34.30} & \cellcolor{graybg}\textbf{0.915} \\
\bottomrule
\end{tabular}
}
\vspace{-20pt}
\end{table*}

\noindent \textbf{Experiments on synthetic Gaussian denoising.}
Following standard protocols~\cite{tassano2020fastdvdnet}, we evaluate AWGN removal on the DAVIS and Set8 benchmarks. As detailed in Tab.~\ref{tab:comparison}, our F2R consistently outperforms unsupervised methods by large margins. On DAVIS, F2R achieves the highest average PSNR of 36.14 dB, surpassing the recent competitor TAP-T~\cite{fu2024temporal} by 0.66 dB and RDRF~\cite{wang2023recurrent} by 0.96 dB. Notably, on the Set8 dataset, F2R (34.30 dB) not only outperforms unsupervised baselines but also surpasses the supervised FloRNN~\cite{li2022unidirectional} by 0.75 dB, narrowing the gap between supervised and unsupervised paradigms. Fig.~\ref{f3} illustrates video denoising examples on DAVIS and Set8 datasets. In the challenging \emph{car-race} scene where even the clean reference is motion-blurred, F2R leverages complementary temporal cues to recover sharper dashboard scale markings and clearer structures than competing methods. More critically, in the dynamic texture region on the pants of the snowboarder, TAP fails to preserve the fabric structure despite incorporating image denoising outputs, indicating suboptimal utilization. Conversely, F2R effectively recovers these high-frequency spatial textures, validating the effectiveness of our strategy.

\begin{figure}[htb]
  \centering
  \includegraphics[width=\linewidth]{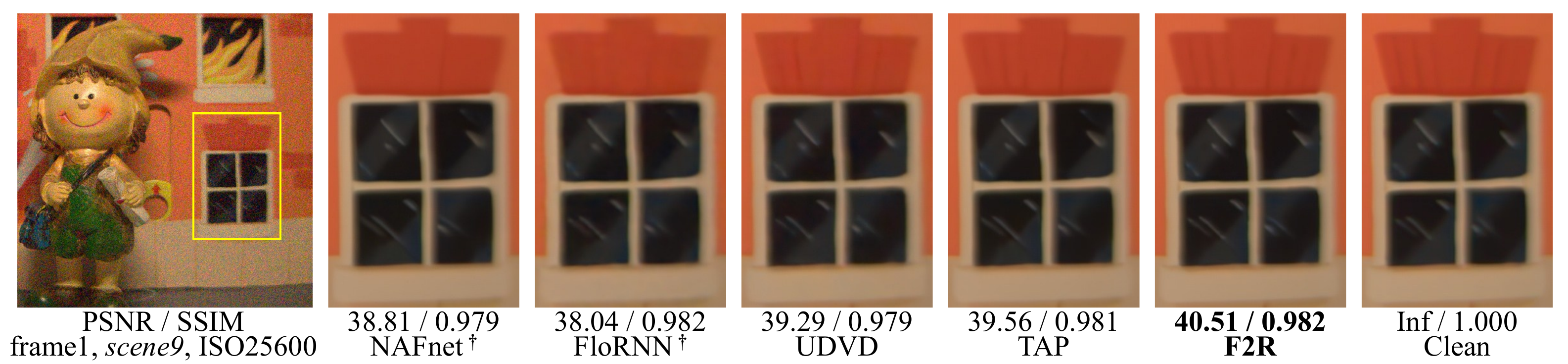}
  \caption{Visual comparison on CRVD indoor dataset. The results have been converted to the sRGB domain with the pretrained ISP provided in \cite{yue2020supervised} for visualization. $\dag$ indicates the supervised method.}
  \label{f4}
\end{figure}

\noindent \textbf{Experiments on real raw video denoising.}
Tab.~\ref{tab:crvd_iso} reports the evaluation results on the CRVD indoor dataset. F2R still outperforms the leading unsupervised scheme TAP~\cite{fu2024temporal} by 0.54 dB and surpasses the supervised competitor FloRNN~\cite{li2022unidirectional}, demonstrating superior generalization in complex real-world noise distributions. To validate the perceptual quality, we provide a visual comparison in Fig.~\ref{f4}. In this challenging low-light indoor scene, preserving the structural integrity of the window panes is critical. Existing methods exhibit characteristic failure modes. The existing Video BSN method, UDVD \cite{zheng2023unsupervised}, fails to retain the sharp edges of the window frame, causing severe blurring. Similarly, the spatial-only baseline NAFNet~\cite{chen2022simple} tends to aggressively suppress high-frequency content, leading to over-smoothed textures. Although TAP \cite{fu2024temporal} leverages temporal information to compensate for spatial deficits, it still fails to fully reconstruct texture details in high-frequency regions. In contrast, F2R effectively reconstructs clear structural lines and authentic wall textures, yielding a result that is visually closest to the clean reference.

\begin{table}[t]
    \renewcommand{\arraystretch}{0.96}
    \centering
    \scriptsize
    \caption{Quantitative comparison on CRVD indoor dataset under different ISO levels. The best results among unsupervised methods are \textbf{bolded}.}
    \label{tab:crvd_iso}
    \vspace{-0.2cm}
    \begin{tabular}{l|l|cccccc}
        \toprule
        & \diagbox[width=8em]{Method}{ISO} & 1600 & 3200 & 6400 & 12800 & 25600 & Average \\
        \midrule
        \multirow{3}{*}{Supervised} 
        & NAFNet \cite{chen2022simple}       & 48.02 & 46.05 & 44.04 & 41.44 & 41.49 & 44.21 \\
        & RViDeNet \cite{yue2020supervised}  & 47.74 & 45.91 & 43.85 & 41.20 & 41.17 & 43.97 \\
        & FloRNN \cite{li2022unidirectional} & 48.81 & 47.05 & 45.09 & 42.63 & 42.19 & 45.15 \\
        \midrule
        \multirow{4}{*}{Unsupervised} 
        & UDVD \cite{zheng2023unsupervised}  & 48.02 & 46.44 & 44.74 & 42.21 & 42.13 & 44.71 \\
        & RDRF \cite{wang2023recurrent}      & 48.38 & 46.86 & 45.24 & 42.72 & 42.25 & 45.09 \\
        & TAP \cite{fu2024temporal}       & 48.85 & 47.03 & 45.11 & 42.44 & 42.33 & 45.15 \\
        
        & \cellcolor{graybg}\textbf{F2R (Ours)}      & \cellcolor{graybg}\textbf{49.41} & \cellcolor{graybg}\textbf{47.52} & \cellcolor{graybg}\textbf{45.66} & \cellcolor{graybg}\textbf{43.03} & \cellcolor{graybg}\textbf{42.82} & \cellcolor{graybg}\textbf{45.69} \\
        \bottomrule
    \end{tabular}
\end{table}

\begin{figure}[htb]
  \centering
  \includegraphics[width=\linewidth]{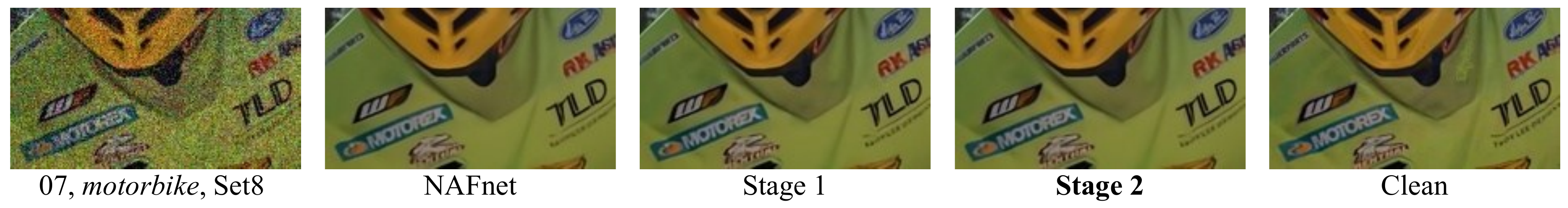}
  \caption{Visual ablation of intermediate stages in F2R. Evaluated on the fast-motion \emph{motorbike} sequence from Set8 with noise level $\sigma$ = 30. \textbf{Stage~2} represents the result of cascaded training of Stage~1 and Stage~2, as denoised by the $f_{\theta'}$.}
  \vspace{-15pt}
  \label{fig:ablation_visual}
\end{figure}

\subsection{Ablation Study}

\textbf{Impact of Individual Stages in F2R.} We validate the necessity of our dual-stage design in Tab.~\ref{tab:ablation_stage}. Compared to the baseline, Stage~1 alone yields significant improvements, confirming its ability to estimate stable temporal residual. Conversely, applying Stage 2 independently fails. Without the temporal anchor from Stage 1, the network simply learns a trivial identity mapping of the image denoising output. This proves that Stage~2 is strictly contingent upon the temporal consistency established by Stage~1. Ultimately, the synergy of both stages achieves peak performance, providing further significant gains beyond Stage~1 alone on the CRVD dataset. As visualized in Fig.~\ref{fig:ablation_visual}, the image denoising output (NAFnet) suffers from severe blur. While Stage~1 stabilizes the background, it omits unique spatial texture, leaving the ``MOTOREX'' and ``RK'' words indistinct due to the intrinsic blindness of the inference process. Stage~2 remedies this via spatial refinement, utilizing the visible center frame to recover sharp, legible text. This validates the effective decomposition of F2R, where Stage~1 secures temporal consistency while Stage~2 refine spatial specificity.

\begin{table}[h]
    \centering
    \scriptsize
    \caption{Impact of individual stages in F2R. We investigate the contribution of Stage 1 and Stage 2. For the Stage 2-only setting, we directly utilize the initial joint inputs to perform re-corruption. Note that the identical scores in the first and third columns are due to rounding.}
    \label{tab:ablation_stage}
    \setlength{\tabcolsep}{11.0pt}
    \begin{tabular}{l|cccc}
        \toprule
        \multirow{1}{*}{Component} & \multicolumn{4}{c}{Methods} \\
        \midrule
        Per-denoised  &\checkmark  & \checkmark  & \checkmark & \checkmark  \\
        Stage 1        &            & \checkmark  &            & \checkmark  \\
        Stage 2        &            &             & \checkmark & \checkmark  \\
        \midrule
        & \multicolumn{4}{c}{\textit{\textbf{DAVIS}} in $\sigma$ = 30} \\
        PSNR (dB)      & 33.47      & 34.36      & 33.47      & \textbf{35.56} \\
        SSIM           & 0.904      & 0.930      & 0.904      & \textbf{0.940} \\
        \midrule
        & \multicolumn{4}{c}{\textit{\textbf{Set8}} in $\sigma$ = 30} \\
        PSNR (dB)      & 31.81      & 32.70      & 31.81      & \textbf{33.82} \\
        SSIM           & 0.869      & 0.903      & 0.869      & \textbf{0.913} \\
        \midrule
        & \multicolumn{4}{c}{\textit{\textbf{CRVD}} in ISO = 25600} \\
        PSNR (dB)      & 41.49      & 41.88      & 41.49      & \textbf{42.82} \\ 
        \bottomrule
    \end{tabular}
\end{table}

\noindent\textbf{Effectiveness of key components in F2R.} We validate our design in Tab. \ref{tab:ablation_component}. The baseline, a standard 4-level U-Net that adopts channel-wise sequence stacking and direct skip connections, suffers from inefficient optimization and feature misalignment arising from implicit motion modeling. Replacing direct skips with FAAMs decouples spatiotemporal features, mitigating alignment issues and obtaining 3.16 dB gain. Most significantly, adopting the joint inputs yields a massive 8.53 dB gain. This confirms that shifting optimization to the residual domain drastically simplifies learning by focusing exclusively on residuals. Ultimately, the synergy of explicit guidance and residual optimization achieves peak performance, validating their strong complementarity.

\begin{table}[h]
    \centering
    \scriptsize
    \caption{Effectiveness of key components. The baseline is a standard 4-level U-Net with channel-wise frame concatenation and direct skip connections. We evaluate replacing direct skips with FAAM ($\mathcal{V}$) and shifting raw inputs to joint inputs ($z$) for residual prediction. Evaluated in Stage 1 on DAVIS ($\sigma=30$).}
    \label{tab:ablation_component}
    \setlength{\tabcolsep}{10.0pt}
    \begin{tabular}{c|cccc}
        \toprule
        \multirow{1}{*}{Component} & \multicolumn{4}{c}{Methods} \\
        \midrule
        Flow-guided ($\mathcal{V}$)&            &\checkmark  &            & \checkmark   \\
        Joint inputs ($z$)         &            &            & \checkmark & \checkmark   \\
        \midrule
        PSNR (dB)                  & 25.06      & 28.22      & 33.59      & \textbf{34.36} \\
        SSIM                       & 0.731      & 0.870      & 0.907      & \textbf{0.930} \\
        \bottomrule
    \end{tabular}
\end{table}

\noindent\textbf{Necessity of Decoupled Alignment Modules.} 
Tab. \ref{tab:ablation_align} verifies that alignment modules adapt to specific objective of each stage. An uniform FAAM (35.37 dB) ensures stability but lacks the sub-pixel precision required for Stage 2 spatial recovery. Conversely, an uniform FDAM (35.53 dB) improves overall alignment but remains suboptimal in Stage 1, as blind estimation lacks the center reference needed to reliably guide complex deformable offsets. By decoupling these modules and utilizing the robust FAAM for Stage 1 blind extraction alongside the precise FDAM for Stage 2 non-blind refinement, F2R achieves peak performance.

\begin{table}[h]
    \centering
    \scriptsize
    \renewcommand{\arraystretch}{0.9}
    \caption{Validation of strategy-specific flow-guided implicit alignment modules on the DAVIS dataset ($\sigma=30$).}
    \label{tab:ablation_align}
    \vspace{-0.2cm}
    \setlength{\tabcolsep}{8.0pt}
    \begin{tabular}{cc|cc}
        \toprule
        Stage 1 & Stage 2 & PSNR (dB) & SSIM \\
        \midrule
        FAAM  & FAAM   & 35.37 & 0.939 \\
        FDAM  & FDAM   & 35.53 & 0.940 \\
        \midrule
        \textbf{FAAM} & \textbf{FDAM} & \textbf{35.56} & \textbf{0.940} \\
        \bottomrule
    \end{tabular}
\end{table}

\noindent\textbf{Influence of Temporal Window Size.} 
To strictly isolate temporal estimating from spatial refinement, we ablate the window size $T \in \{3, 5, 7, 9, 11\}$ exclusively on the Stage 1 intermediate output ($\hat{x}_{s1}$) using synthetic datasets ($\sigma=30$). As Tab.~\ref{tab:ablation_temporal} shows, while increasing $T$ initially boosts performance, it quickly exhibits diminishing returns. Expanding from $T=9$ to $T=11$ yields negligible gains (e.g., +0.03 dB on DAVIS) while significantly exacerbating computational overhead and the risk of long-range flow alignment artifacts. Thus, $T=9$ serves as the optimal trade-off for synthetic datasets. For the real-world CRVD dataset, we utilize $T=7$, naturally constrained by its maximum captured sequence length.

\noindent\textbf{Fair Comparison under Identical Temporal Window.}
To ensure a fair comparison, we evaluate F2R under the strict 5-frame setting ($T=5$) adopted by baselines such as FastDVDNet \cite{tassano2020fastdvdnet}, UDVD \cite{zheng2023unsupervised}, and TAP-T \cite{fu2024temporal}. As reported in Tab.~\ref{tab:fair_comparison}, F2R achieves 35.19 dB on the DAVIS dataset ($\sigma=30$) in this restricted context. This outperforms the leading method, TAP-T, by 0.35 dB, confirming the superiority of F2R.

\begin{table}[h]
    \centering
    \scriptsize
    \caption{Ablation on the temporal window size $T$. Results report PSNR (dB) on the Stage 1 intermediate output using synthetic datasets ($\sigma=30$). Values in parentheses indicate the marginal gain over the previous $T$.}
    \label{tab:ablation_temporal}
    \setlength{\tabcolsep}{6.0pt}
    \begin{tabular}{c|ccccc}
        \toprule
        \multirow{2}{*}{Dataset} & \multicolumn{5}{c}{Temporal Window Size $T$} \\
        \cmidrule{2-6}
        & $T=3$ & $T=5$ & $T=7$ & $T=9$ & $T=11$ \\
        \midrule
        DAVIS & 33.96 & 34.23 (+0.27) & 34.31 (+0.08) & \textbf{34.36 (+0.05)} & 34.39 (+0.03) \\
        Set8  & 32.30 & 32.56 (+0.26) & 32.65 (+0.09) & \textbf{32.70 (+0.05)} & 32.74 (+0.04) \\
        \bottomrule
    \end{tabular}
\end{table}

\begin{table}[h]
    \centering
    \scriptsize
    \caption{Fair comparison under an identical temporal window ($T=5$). Evaluated on the DAVIS dataset under the noise level $\sigma=30$. All methods strictly utilize 5 input frames. Best results are highlighted in \textbf{bolded}.}
    \label{tab:fair_comparison}
    \setlength{\tabcolsep}{10.0pt}
    \begin{tabular}{l|cc|cc}
        \toprule
        Method & Type & Frames ($T$) & PSNR (dB) & SSIM \\
        \midrule
        FastDVDNet \cite{tassano2020fastdvdnet} & Supervised & 5 & 34.04 & 0.917 \\
        UDVD \cite{zheng2023unsupervised}       & Self-supervised & 5 & 34.09 & 0.918 \\
        TAP-T \cite{fu2024temporal}             & Self-supervised & 5 & 34.84 & 0.926 \\
        \midrule
        \textbf{F2R (Ours)}                     & Self-supervised & 5 & \textbf{35.19} & \textbf{0.934} \\
        \bottomrule
    \end{tabular}
\end{table}

\section{Conclusion}
In this paper, we propose F2R, a framework empowered by a \emph{spatiotemporal decoupling} strategy, designed to address the critical bottleneck of severed \emph{spatiotemporal correlations} in existing Video BSNs. Fundamentally, F2R operates through two tailored self-supervised strategies. It first employs a frame-wise blind strategy to robustly estimate a temporally consistent anchor. Subsequently, a recorruption strategy leverages this anchor and reintroduces the excluded frame to safely recover the high-frequency spatial residuals discarded by the image denoiser. This decoupled design ensures that the recovered spatial specificity strictly maintains the established temporal stability, effectively reconstructing the \emph{spatiotemporal correlations}. Comprehensive experiments on sRGB and raw video datasets verify that F2R achieves state-of-the-art performance.

\subsubsection*{Acknowledgements}
This work was supported by the National Key Research and Development Program of China (2023YFF0713300).

%
%
\bibliographystyle{splncs04}
\bibliography{main}
\end{document}